\documentclass[conference]{IEEEtran}

\usepackage{amsmath}
\usepackage{hyperref}
\usepackage{url}
\usepackage{graphicx}
\usepackage{subfig}
\usepackage[table,xcdraw]{xcolor}
\usepackage{caption}
\usepackage{placeins}

\DeclareMathOperator{\argmin}{\text{arg\,min}}

\newcommand{\eg}{e.\,g.~}
\newcommand{\ie}{i.\,e.~}
\newcommand{\etal}{\emph{et al.~}}

\renewcommand{\vec}[1]{\mathbf{#1}}
\newcommand{\mat}[1]{\mathbf{#1}}

\newcommand{\Sref}[1]{{Sec.~\ref{#1}}}
\newcommand{\Eref}[1]{{Eq.~\ref{#1}}}
\newcommand{\Fref}[1]{{Fig.~\ref{#1}}}
\newcommand{\Tref}[1]{{Table~\ref{#1}}}

\newcommand{\tf}[2]{\ensuremath{^{\text{#1}}\mat{T}^{}_{\text{#2}}}}
\newcommand{\tfinv}[2]{\ensuremath{^{\text{#1}}\mat{T}^{-1}_{\text{#2}}}}

\DeclareRobustCommand*{\IEEEauthorrefmark}[1]{%
	\raisebox{0pt}[0pt][0pt]{\textsuperscript{\footnotesize #1}}%
}

\begin{document}
%
\title{Augmented Reality-based Feedback for Technician-in-the-loop C-arm Repositioning}

\author{\IEEEauthorblockN{Mathias Unberath\IEEEauthorrefmark{1,$\ast$}, Javad Fotouhi\IEEEauthorrefmark{1,$\ast$}, Jonas Hajek\IEEEauthorrefmark{1,2,$\ast$}, Andreas Maier\IEEEauthorrefmark{2}, Greg Osgood\IEEEauthorrefmark{3}, Russell Taylor\IEEEauthorrefmark{4},\\ Mehran Armand\IEEEauthorrefmark{5}, and Nassir Navab\IEEEauthorrefmark{1}}

\IEEEauthorblockA{\IEEEauthorrefmark{1}Computer Aided Medical Procedures, Johns Hopkins University}
\IEEEauthorblockA{\IEEEauthorrefmark{2}Pattern Recognition Laboratory, FAU Erlangen-Nuremberg}
\IEEEauthorblockA{\IEEEauthorrefmark{3}Orthopaedic Trauma, Johns Hopkins University}
\IEEEauthorblockA{\IEEEauthorrefmark{4}Laboratory for Computational Sensing and Robotics, Johns Hopkins University}
\IEEEauthorblockA{\IEEEauthorrefmark{5}Applied Physics Laboratory, Johns Hopkins University}
\IEEEauthorblockA{\IEEEauthorrefmark{$\ast$}These authors are considered joint first authors.}
\IEEEauthorblockA{Email: unberath@jhu.edu}}
\maketitle

\begin{abstract}
Interventional C-arm imaging is crucial to percutaneous orthopedic procedures as it enables the surgeon to monitor the progress of surgery on the anatomy level. Minimally invasive interventions require repeated acquisition of X-ray images from different anatomical views to verify tool placement. Achieving and reproducing these views often comes at the cost of increased surgical time and radiation dose to both patient and staff. This work proposes a marker-free "technician-in-the-loop" Augmented Reality (AR) solution for C-arm repositioning. The X-ray technician operating the C-arm interventionally is equipped with a head-mounted display capable of recording desired C-arm poses in 3D via an integrated infrared sensor. For C-arm repositioning to a particular target view, the recorded C-arm pose is restored as a virtual object and visualized in an AR environment, serving as a perceptual reference for the technician.\\ 
We conduct experiments in a setting simulating orthopedic trauma surgery. Our proof-of-principle findings indicate that the proposed system can decrease the 2.76 X-ray images required per desired view down to zero, suggesting substantial reductions of radiation dose during C-arm repositioning. The proposed AR solution is a first step towards facilitating communication between the surgeon and the surgical staff, improving the quality of surgical image acquisition, and enabling context-aware guidance for surgery rooms of the future. The concept of technician-in-the-loop design will become relevant to various interventions considering the expected advancements of sensing and wearable computing in the near future.
\end{abstract}

\section{Introduction}
\label{sec:intro}
Percutaneous approaches are the current clinical standard for internal fixation of many skeletal fractures, including pelvic trauma. This type of minimally invasive surgery is enabled by C-arm X-ray imaging systems that intra-operatively supply projective 2D images of the 3D surgical scene, \ie tools and anatomy. Appropriate placement of implants is crucial for satisfactory outcome~\cite{korovessis2000medium,mardanpour2013outcome}, but verifying acceptable progress interventionally is challenging. This is because it requires the mental mapping of desired screw trajectories to the fractured anatomy in 3D based on 2D X-ray images acquired from different view points~\cite{tucker2018towards,andress2018fly,hajek2018closing}. To alleviate the associated challenges, surgeons are trained to use well defined X-ray views specific to the current task, such as the inlet or obturator oblique view of the pelvis~\cite{wheeless1996ortho,AOsurgical}. Achieving these views, however, is not straight forward in practice due to multiple reasons. First, the desired views are usually difficult to obtain since the position of the internal anatomy is not obvious from the outside. Second, C-arm systems most commonly used today are non-robotic but have many degrees of freedom. Consequently, when trying to achieve a particular view, the X-ray technician operating the C-arm positions the device by repeated trial-and-error. Doing so increases the radiation dose to patient and surgical team. In surgical workflows where the C-arm has to be moved out of the way to ease access to the patient (as is the case in pelvis fixation), the problem of increased dose during so-called ''fluoro hunting''~\cite{desilva2018VirtualFluoro} is further amplified. The above reasoning suggests that a computer-assisted solution that aids the X-ray technician in finding the desired view has great potential in reducing X-ray dose to patient and surgical staff.

Most previous work have focused on digitally rendering X-ray images from CT data rather than physically acquiring them. \cite{bott2011informatics,gong2014cost,stefan2017mixed} use ''virtual fluoroscopy'' to improve training of X-ray technicians and surgeons, while~\cite{klein2007interactive,dressel2010intraoperative} and~\cite{desilva2018VirtualFluoro} generate digitally rendered radiographs intra-operatively from pre-operative CT. Doing so requires 3D/2D registration of the CT volume to the patient and tracking of the C-arm, which is achieved using an additional RGB camera or C-arm encoders, respectively. A complementary method most similar to the approach discussed here uses an external outside-in tracking system that accurately tracks an optical marker on the C-arm to verify accurate repositioning~\cite{matthews2007navigating}. All the above approaches successfully reduce radiation dose due to C-arm repositioning, however, they make strong assumptions on the surgical environment by requiring pre-operative CT, an encoded C-arm, or external tracking systems. 

In~\cite{fallavollita2014desired}, a user interface concept is introduced for navigating and repositioning angiographic C-arms. First, the surgeon identifies the desired imaging outcome based on radiographs simulated from pre-operative CTA images on a tablet PC system. The 6 \textit{degree-of-freedom} pose of the C-arm scanner is automatically estimated by using this planning information, the registration between the patient and the scanner, and the inverse kinematics of the C-arm. Consequently, this solution provides a transparent interface to the control of the imaging device. However, a major challenge associated to it is the offline planning stage which prohibits its application and usefulness for percutaneous orthopedic interventions considering their ergonomics and dynamic workflow. 

In this work, we propose a technician-in-the-loop solution to C-arm repositioning during orthopedic surgery in unprepared operating theaters. This is achieved by equipping the X-ray technician with an optical see-through head-mounted display (OST HMD) that tracks itself within its environment. Then, once the C-arm is positioned appropriately, the 3D point cloud of the C-arm is stored. When a particular pose must be restored, the respective 3D scene is visualized to the technician in an Augmented Reality (AR) environment, providing intuitive feedback on C-arm positioning in 3D. 

\begin{figure}\centering
	\includegraphics[width=1.0\linewidth]{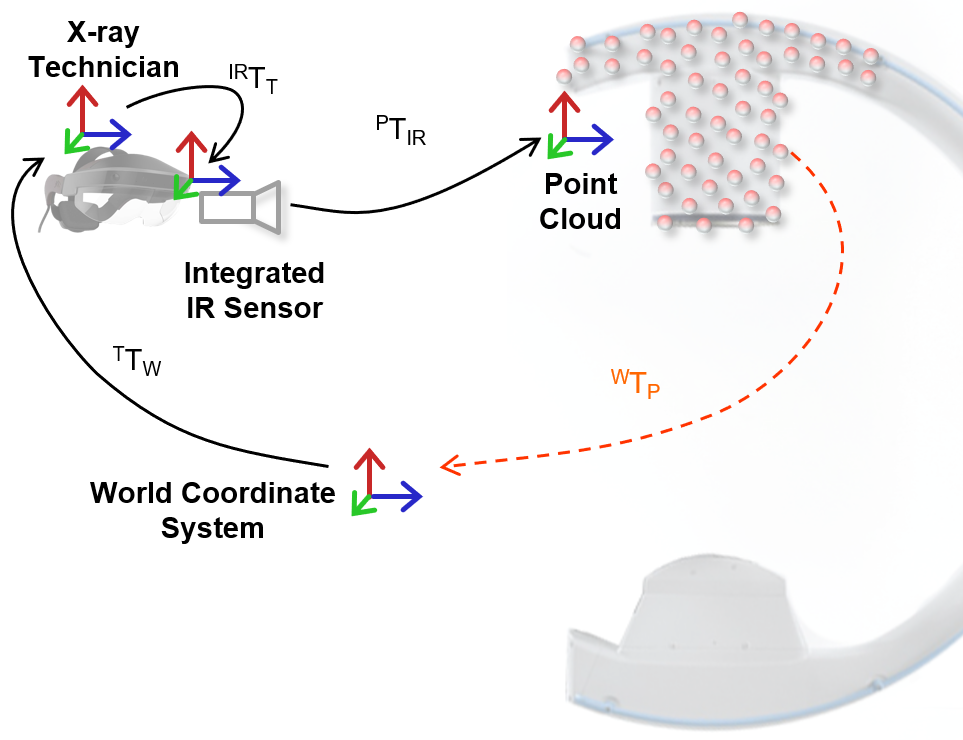}
	\caption{Spatial relations that must be estimated dynamically to enable the proposed AR environment. Transformations shown in black are estimated directly while transformations shown in orange are derived.}
	\label{fig:transforms}
\end{figure}

\section{Methods}
\label{sec:methods}
Similarly to the AR environment delivered by the camera-augmented C-arm~\cite{tucker2018towards}, the proposed solution for C-arm repositioning does not actively track the device to be positioned but intuitively visualizes spatial relations and thus improves user performance. To this end, several transformations need to be estimated dynamically. These transforms are illustrated in \Fref{fig:transforms} and their estimation is discussed in the remainder of this section. 

\begin{figure*}[tb!]\centering
	\includegraphics[width=1.0\linewidth]{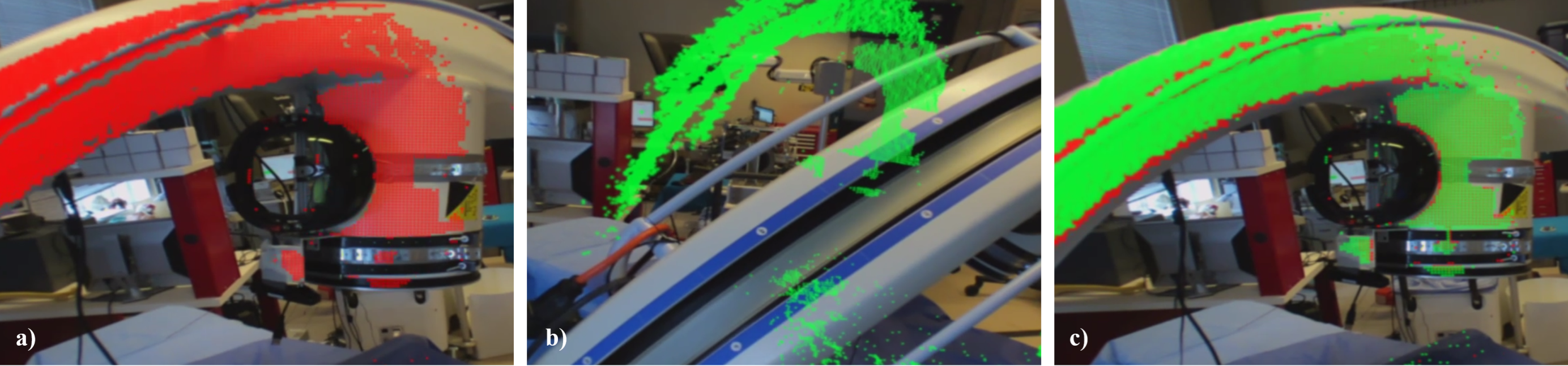}
	\caption{All images are shown from the X-ray technician's point of view. In a) the live 3D point cloud computed from the infrared depth image is displayed in red. This point cloud is then saved for re-use.  In b) the C-arm has been moved to a different pose; the previously saved point cloud is visualized in green and serves as a reference to achieve the previous pose. After successful repositioning of the C-arm shown in c), the saved and current point clouds shown in green and red, respectively, coincide. This means that the C-arm has been repositioned appropriately.}
	\label{fig:technician_pov}
\end{figure*}

\subsection{Localization in the Operating Theater}
\paragraph{Tracking the Technician} The central mechanism of the proposed system is the ability to store the 3D appearance of a C-arm configuration (\ie the 3D point cloud as shown in \Fref{fig:technician_pov}) at the respective position in 3D space; and to recreate it in an AR environment. We use the Simultaneous Localization and Mapping (SLAM) capabilities of the OST HMD to dynamically calibrate the headset, and thus the technician, to its environment. Using image features obtained via depth sensors or stereo cameras, SLAM incrementally constructs a spatial map of the environment and localizes the sensing device therein~\cite{endres2012evaluation}. In particular, SLAM solves
\begin{equation}
\tf{W}{T}(t) = \underset{^{\text{W}}\mat{\hat{T}}_{\text{T}}}{\argmin}\ \mathrm{d}\!\left[  \mat{f}_{\text{W}}\left(\mat{P}\, ^{\text{W}}\mat{\hat{T}}_{\text{T}}(t)\mat{x}_{\text{T}}(t)\right), \mat{f}_{\text{T}}(t)   \right],
\label{eq:slam}
\end{equation}
where $\tf{W}{T}(t)$ is the desired pose of the technician relative to an arbitrary but static world coordinate system at time $t$, $\mat{f}_{\text{T}}(t)$ are image features at that time, $\mat{x}_{\text{T}}(t)$ are the 3D locations of these feature, $\mat{P}$ is the projection operator, and $d[\cdot,\cdot]$ is the feature similarity to be optimized~\cite{endres2012evaluation,hajek2018closing}.

\paragraph{Sensing the C-arm Position} 
In contrast to the technician, the C-arm is not tracked explicitly but only imaged in 3D using an infrared depth camera integrated in the HMD worn by the technician. Since the X-ray technician works in reasonable proximity of the C-arm, the infrared sensor will constantly observe a large area of the C-arm's surface (cf. \Fref{fig:technician_pov}). Once the C-arm has been moved to the desired location, the technician can save the C-arm position by either voice command (\eg "Save Position 1") or by pressing a button on a hand-held remote control. Here, saving the position refers to saving the current point cloud $\{\vec{x}^\text{P}_i\,|\,i=1,\dots,N\}$ relative to the world coordinate system via
\begin{equation}
\tf{W}{P}(t_0) = \tfinv{W}{T}(t_0) \cdot \tfinv{IR}{T} \cdot \tfinv{P}{IR} \cdot \vec{x}^\text{P},
\label{eq:twp}
\end{equation}
where $t_0$ is the time of voice command, $\tf{IR}{T}$ is the HMD-specific transformation from tracking module of the HMD to its infrared camera, and $\tf{P}{IR}$ is the mapping from infrared sensor to metric 3D points $\vec{x}^\text{P}$. 

\subsection{Guidance by Visualization}
\label{subsec:guidance}
The process of saving C-arm positions is repeated for every desired X-ray view such that point clouds of the C-arm device in every pose are available. During the procedure when previous C-arm views have to be re-produced, the X-ray technician requests visualization of the desired position via voice command (''Show Position 1''). Since the point cloud $\{\vec{x}^\text{P}_{j,i}\,|\,i=1,\dots,N_j\}$ of C-arm position $j$ is stored relative to the world coordinate frame, it can be visualized to the X-ray technician in an AR environment at position 
\begin{equation}
\vec{x}^\text{T}_{j,i}(t) =  \tf{T}{W}(t) \cdot \underbrace{\tf{W}{P}(t^j_0) \cdot \vec{x}^\text{P}_{j,i}(t_0^j)}_{\vec{x}^\text{W}_{j,i}}\,,
\end{equation}
where $t^j_0$ denotes the time of calibration of view $j$, $t$ is the current time, $\vec{x}^\text{W}_{j,i}$ is the $i^\text{th}$ point in point cloud $j$ in the world coordinate frame, and $\tf{W}{P}$ is computed according to \Eref{eq:twp}. An example of the AR environment during visualization of a representative point cloud is provided in \Fref{fig:technician_pov}b).\\
In contrast to previous approaches, there is no explicit guidance but intuitive 3D visualization of the desired position. The X-ray technician adjusts the position of the C-arm using all available degrees of freedom (axial, orbital and swivel rotation in addition to base and gantry translations) such that the surface of the real C-arm perfectly matches the virtual point cloud. The live point cloud $\{\vec{x}^\text{P}{i}(t)\,|\,i=1,\dots,N(t)\}$ can be toggled on or off for additional virtual-on-virtual assessment (see \Fref{fig:technician_pov}).

\subsection{Experiments and Study}
\begin{figure}[tb!]\centering
	\includegraphics[width=\linewidth]{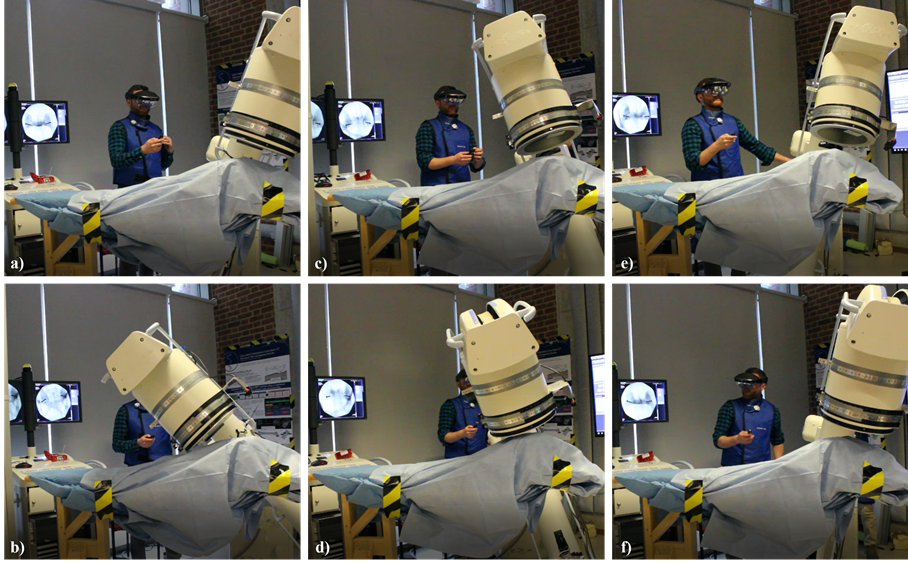}
	\caption{The X-ray technician operating the C-arm during the experiment. Typical angulations for pelvic trauma surgery were selected according to~\cite{wheeless1996ortho}. We show inlet and outlet views in a) and b), caudal oblique views in c) and d), and cranial oblique views in e) and f), respectively.}
	\label{fig:carm_poses}
\end{figure}

To test the described system we setup an experiment mimicking pelvic trauma surgery using an anthropomorphic Sawbones pelvis phantom (Sawbones, Vashon, WA). The phantom was completely covered with surgical drape and had metallic markers attached to define keypoints for evaluation. During this study, the C-arm was operated by a board-certified X-ray technician, who usually operates C-arm imaging systems during surgery. During the experiment and for every run, the X-ray technician was asked to: First, move the system into two clinically relevant C-arm poses, \eg inlet and outlet view; second, retract the C-arm and reset to neutral position; and third, accurately reproduce the two previously defined C-arm positions using the conventional method (\ie no assistance) and the proposed AR environment. Representative angulations of the C-arm are shown in \Fref{fig:carm_poses}. For direct quantitative comparison between C-arm poses, an infrared optical marker was rigidly attached to the gantry of the C-arm and tracked using an external tracking camera, namely a Polaris Spectra (Northern Digital Inc., Shelburne, VT).
The workflow for one run was as follows:
\begin{itemize}
	\item[Step 1:] Define two target C-arm poses, save X-ray images, point cloud using HMD, and C-arm position using external tracker
	\item[Step 2:] Retract C-arm from scene and set in neutral position
	\item[Step 3:] Restore target views
	\begin{itemize}
		\item Conventional: Store all X-ray views required for repositioning, and final C-arm position using external marker
		\item Proposed: Store final C-arm position, and one X-ray image for evaluation
	\end{itemize}
	\item[Step 4:] Repeat Step 3 with other method (conventional/proposed)
\end{itemize}
We designed four runs covering a total of six different poses:
\begin{itemize}
	\item[Run 1:] Inlet/outlet
	\item[Run 2:] Cranial oblique / Caudal oblique
	\item[Run 3:] Cranial oblique / Caudal oblique (opposing)
	\item[Run 4:] Inlet / outlet
\end{itemize}
To avoid training bias, we alternate the order in which conventional and proposed approach are utilized for every run.\\
For quantitative evaluation, we report the mean Euclidean and angular difference of final C-arm poses compared to the target pose as measured by the external tracker. Further, we manually annotate the keypoint locations in all X-ray images and compute the average projection domain displacement using the first and final X-ray for the conventional approach, and using the verification X-ray for the proposed method. Finally, we record the total number of X-rays used during the conventional repositioning.

\begin{table}
	\centering
	\caption{C-arm pose differences as per infrared marker tracking.}
	\label{tab:posetracker}
	\begin{tabular}{l c c}
		\rowcolor[HTML]{EFF4FB} 
		\multicolumn{1}{c}{\cellcolor[HTML]{EFF4FB}} & \multicolumn{1}{c}{\cellcolor[HTML]{EFF4FB}\textbf{Proposed}} & \multicolumn{1}{c}{\cellcolor[HTML]{EFF4FB}\textbf{Conventional}} \\ \hline
		\multicolumn{1}{l|}{\textbf{Mean Distance $\pm$ SD}} & \multicolumn{1}{c|}{51.6 $\pm$ 19.2\,mm} & 16.7 $\pm$ 6.3\,mm \\
		\multicolumn{1}{l|}{\textbf{Angle $\pm$ SD}} & \multicolumn{1}{c|}{1.54 $\pm$ 0.92$^\circ$} & 1.23 $\pm $ 0.45$^\circ$
	\end{tabular}
\end{table}

\section{Results}
We have omitted Run 3 from the quantitative evaluation, since the X-ray technician erroneously restored the cranial oblique view as requested in Run 2. This is because inclusion of this run would strongly bias the quantitative results of the conventional approach, and thus, positively bias the assessment of the proposed AR environment. In clinical practice, such errors unnecessarily increase the dose to patient and surgical staff but could be avoided using the proposed system.\\ 
Differences between target and restored C-arm pose as per the external tracker are provided in \Tref{tab:posetracker} for the proposed and conventional approach, respectively. We state residuals averaged over all C-arm poses separately for translation and rotation and compute the mean Euclidean displacement and angular deviation of the reference marker, respectively. While the orientation of the C-arm is equally well restored in both proposed and conventional approach, the positional error is larger for the proposed method. This observation will be discussed in the following section.\\
In addition, we state the average displacement of projection domain keypoints relative to the target X-ray images. We evaluate this error for the verification X-ray images after C-arm repositioning with the proposed method, and for the initial and refined X-ray images acquired in the conventional approach. The values stated in \Tref{tab:firsttry} reflect the mean pixel displacement over all poses and keypoints. 3 to 4 keypoints were used per image, depending on the field of view determined by the C-arm pose. Based on this projection domain metric, the proposed method outperforms C-arm repositioning based on user recollection, \ie the conventional approach before iterative refinement. However, when X-ray images are acquired to verify and adjust the C-arm pose, the conventional approach substantially outperforms the proposed system.\\
Finally, we report the number of X-ray images acquired during C-arm repositioning. With the conventional method, a total of 16 X-rays where required to restore 6 poses yielding, on average, 2.76 X-rays per C-arm position. Using the proposed approach, the number of acquired images for C-arm pose restoration drops to zero, since our experiment did not allow for iterative refinement when the proposed technology was used.  

\begin{table}
	\centering
	\caption{Projection domain keypoint displacement in pixels (px).}
	\label{tab:firsttry}
	\begin{tabular}{ccc}
		\rowcolor[HTML]{EFF4FB} 
		\textbf{Proposed} & \textbf{Conventional on first try} & \textbf{Conventional} \\ \hline
		\multicolumn{1}{c|}{210 $\pm$ 105\,px} & \multicolumn{1}{c|}{257 $\pm$ 171 px} & 68 $\pm$ 36 px
	\end{tabular}
\end{table}

\section{Discussion}
Our results suggest that substantial dose reductions are possible with the proposed AR system. Once the desired views have been identified and stored, C-arm repositioning can be achieved with clinically acceptable accuracy without any further X-ray acquisitions. At the same time, our results reveal that further research on improving tracking accuracy and perceptual quality will be required for X-ray technicians to not only save dose but also deliver improved performance. 
The experimental design described here is limited since only a single X-ray technician and four runs were considered. We understand the reported experiments as an exploratory study designed to reveal shortcomings of the current prototype. Based upon these very preliminary results, we envision necessary refinements of the system that address the current challenges discussed in greater detail below.\\
In contrast to previous methods~\cite{klein2007interactive,dressel2010intraoperative,desilva2018VirtualFluoro}, our approach can be directly deployed in the operating theater without any preparation of the environment or assumptions on the procedure. This translates to two immediate benefits: First, our method does not require pre- or intra-operative 3D imaging, and therefore, circumvents intra-operative 3D/2D registration, a major challenge in clinical deployment~\cite{bier2018x,fotouhi2017pose}. Second, there is no need for additional markers and external trackers as in Matthews \etal~\cite{matthews2007navigating}, or access to internal encoders of the C-arm as in De Silva \etal~\cite{desilva2018VirtualFluoro}. While internal encoders can be considered more elegant than external trackers, they are not yet widely available in mobile C-arms since these systems are usually non-robotic, and therefore, do not require encoding. In addition, the most recent C-arms that are commercially available, such as the Ziehm Vision RFD 3D (Ziehm Imaging, Vienna, Austria) or the Siemens Cios Alpha (Siemens Healthineers, Erlangen, Germany) have at most 4 robotized axes~\cite{desilva2018VirtualFluoro,ziehmimaging}, suggesting that not all of the required 6 degrees of freedom can be monitored.\\
The proposed system does not currently provide quantitative feedback on how well a previously achieved pose was restored, but relies on the user's assessment. The current prototype, however, is capable of simultaneously displaying stored and live point clouds, as described in \Sref{subsec:guidance}. A natural next step would be to use quantitative methods such as the iterative closest point (ICP) algorithm~\cite{besl1992method} to provide rigorous feedback on both the accuracy of alignment and the required adjustments. We strongly believe that such information would substantially improve the performance of the proposed system since current verification is solely based on perception. Our results suggest that relying on perception works well for restoring C-arm orientation but does not perform well for restoring position. This limitation partly arises from the disadvantages and challenges with current technology, particularly because of two reasons: First, there is no interaction between real and virtual object, such as shadow or occlusion. Second, available hardware, such as the Microsoft HoloLens or the Meta 2, will render virtual content in a fixed focal plane, irrespective of the virtual objects position. Consequently, virtual and real content may not be in focus simultaneously despite occupying the exact same physical space~\cite{andress2018fly,hajek2018closing}. 
Integrating quantitative feedback, however, will require optimized implementations of rendering and alignment to deliver a pleasant user experience without substantial lag; a challenge that already arises for pure visualization due to the immense computational load associated with real-time SLAM.\\
In addition to shortcomings regarding perception, the performance of our prototype system is further compromised by the SLAM-tracking performance of purchasable hardware. Our prototype was materialized using the Meta 2 (Meta, San Mateo, CA) since it was the only HMD that provided developer access to the infrared depth sensor at the time of implementation. Unfortunately, we have found the SLAM-tracking provided by the Meta 2 to be inferior to the HoloLens with respect to both lag and accuracy. In addition, the Meta 2 is cable-bound, which limits its appropriateness in highly dynamic environments such as operating theaters. While previous work suggests that the display quality of current HMDs may be sufficient for intra-operative visualization~\cite{qian2017towards,deib2018image,qian2017comparison}, the accuracy and reliability of vision-based SLAM seems yet insufficient to warrant immediate clinical deployment~\cite{andress2018fly,hajek2018closing}. While incorporating external tracking may be a solution~\cite{chen2015development}, we believe that this prerequisite will inhibit wide acceptance, as was observed with previous navigation techniques. Consequently, developing OST HMDs specifically designed to meet clinical needs, particularly regarding perceptual quality and tracking accuracy, will be of critical importance to bring medical AR into the operating room. 

\section{Conclusion}
We have proposed a technician-in-the-loop solution to C-arm repositioning during fluoroscopy-guided procedures. Our system stores 3D representations of the desired C-arm views using real-time 3D sensing via infrared depth cameras that are then stored. When a previously achieved pose needs to be restored, the corresponding 3D scene is displayed to the technician in an OST HMD-based AR environment. Achieving the target view then requires alignment of the real C-arm gantry with the virtual model thereof. In our proof-of-principle experiments we have found that use of our system 1) is associated with a reduction in X-ray dose, and 2) may prevent operator errors, such as restoring the wrong view. We have found that relying on perception as the only performance feedback mechanism is challenging with current HMD hardware, suggesting that future work should investigate possibilities to provide quantitative feedback on C-arm operator performance in real time.

\FloatBarrier



\end{document}